# Timeseries-aware Uncertainty Wrappers for Uncertainty Quantification of Information-Fusion-Enhanced AI Models based on Machine Learning

Janek Groß
*Data Science*
*Fraunhofer IESE*
Kaiserslautern, Germany
janek.gross@iese.fraunhofer.de

Michael Kläs
*Data Science*
*Fraunhofer IESE*
Kaiserslautern, Germany
michael.klaes@iese.fraunhofer.de

Lisa Jöckel
*Data Science*
*Fraunhofer IESE*
Kaiserslautern, Germany
lisa.joeckel@iese.fraunhofer.de

Pascal Gerber
*Data Science*
*Fraunhofer IESE*
Kaiserslautern, Germany
pascal.gerber@iese.fraunhofer.de

*Abstract*— As the use of Artificial Intelligence (AI) components in cyber-physical systems is becoming more common, the need for reliable system architectures arises. While data-driven models excel at perception tasks, model outcomes are usually not dependable enough for safety-critical applications. In this work, we present a timeseries-aware uncertainty wrapper for dependable uncertainty estimates on timeseries data. The uncertainty wrapper is applied in combination with information fusion over successive model predictions in time. The application of the uncertainty wrapper is demonstrated with a traffic sign recognition use case. We show that it is possible to increase model accuracy through information fusion and additionally increase the quality of uncertainty estimates through timeseries-aware input quality features.

## I. INTRODUCTION

Due to generally increasing demands on the behavioral complexity and autonomy of cyber-physical systems (CPS), the number of CPS containing models based on data-driven AI approaches such as machine learning (ML) is also increasing. Particularly for perception tasks, these models, which include deep neural networks, enable functionality that traditionally developed software is currently not able to provide with compatible levels of performance [1].

However, ML models usually cannot be guaranteed by design to provide the intended outcome for all inputs, leading to outputs with inherent uncertainty [2]. Moreover, most common software verification and validation (V&V) methods are not directly applicable to ML models [3] and many advanced V&V methods, such as model-based approaches [4] [5], appear hardly applicable. Although the literature also discusses ML-specific methods [6] and collaboration between the software testing and the data science communities is considered fruitful [7], the application of ML-specific V&V methods during CPS development still appears to be limited.

One way to reduce the probability of critical failures caused by AI components in CPS is to rely not only on design-time V&V such as statistical testing but also use runtime mechanisms. A common example of the latter is monitoring the ML model during operation and detecting outcomes with high uncertainty to either overwrite these outcomes (cf. simplex pattern [8] [9]) or take some other countermeasures on the level of the system that employs the AI component to avoid unsafe behavior [10]. Obviously, the applicability of such runtime V&V depends on the availability of *dependable and situation-aware uncertainty estimates* for the outcomes of the ML model, as provided, e.g., by uncertainty wrappers [11].

Despite advances in AI design and related V&V, the error rates of ML models have still not reached levels that are low enough for many interesting applications of CPS. Thus, an increasing number of publications investigate how to combine perception outcomes using *information fusion* approaches to obtain results with reduced error probability. Outcomes to be fused can be obtained using data from different sources, e.g., different sensors and ML models, or data collected by repeated measurements, e.g., a timeseries of sensor values.

While a variety of information fusion approaches exist [12] [13], new challenges arise when it comes to determining the uncertainty of the fused information at runtime since the combined outcomes are usually not independent in terms of uncertainty.

In this paper, we investigate means to provide dependable uncertainty estimates for runtime V&V in the context of information fusion based on repeated measurements. More specifically, we address the task of image-based *traffic sign recognition* (TSR) considering individual outcomes and uncertainty estimates for each input image of a series.

The main *contributions* can be summarized as follows:

(1) Using the example of TSR, we show that even a simple information fusion approach applied to timeseries data can significantly improve ML model outcomes.

(2) We introduce timeseries-aware uncertainty wrappers (taUWs) as an extension of the existing uncertainty wrapper approach to provide dependable uncertainty estimates for outcomes based on information fusion.

(3) We investigate the performance of taUWs for the TSR use case and compare the results with existing

proposals on how to obtain uncertainty estimates for outcomes based on information fusion.

(4) We conducted a sensitivity analysis to assess which timeseries-related quality factors considered by our approach are most relevant for providing dependable uncertainty estimates.

We begin by reviewing the state of the art for uncertainty estimation in ML and the challenges posed by information fusion on timeseries data (Section 2). Next, we propose a new approach to timeseries-aware uncertainty estimates (Section 3), outline our research questions and describe the study we conducted (Section 4), and then present and discuss the study results (Section 5). Finally, we summarize our conclusions and provide an outlook on future work (Section 6).

## II. BACKGROUND AND RELATED WORK

In this paper, we make a proposal to extend the uncertainty wrapper approach [11] to build *timeseries-aware Uncertainty Wrappers* (taUW), which provide dependable uncertainty estimates also in the context of information fusion based on multiple measurements from one source. Thus, this section first summarizes the concept of uncertainty estimation and introduces uncertainty wrappers as a model-agnostic method for dependable uncertainty estimation. Then we give an overview of information fusion and related work in uncertainty fusion, reviewing existing proposals that might be alternatives to the taUW approach.

***Uncertainty Estimation.*** We cannot expect the outcomes of *data-driven models* (DDMs) [14], which include ML models, to always be correct. Uncertainty estimates provided at runtime for a DDM indicate how much one can rely on the DDM outcome in the current situation [15]. There are various notions of uncertainty, yet in the following we refer to uncertainty as the probability $u$ of a specific failure mode. Certainty is defined as $c = 1 - u$, respectively. Since uncertainty results can only be interpreted if we know the failure mode considered, in our example we stick with the simple failure mode that the outcome of a TSR model for a given input does not correspond to the actual class of the depicted traffic sign (e.g., for an image of a speed limit sign, the DDM provides 'stop_sign' as the outcome).

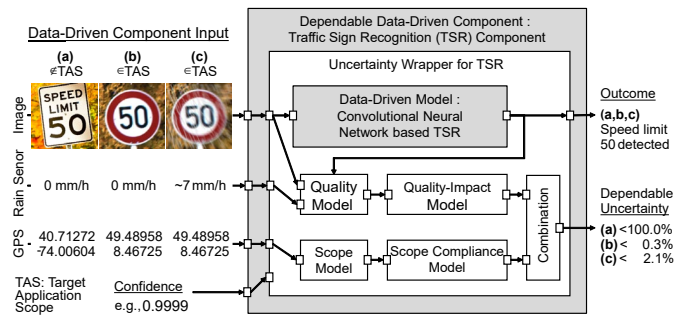

Fig. 1. Uncertainty wrapper pattern [16]

***Uncertainty Wrappers (UW)*** represent an outside-model approach for uncertainty estimation that is model-agnostic. Here, the encapsulated DDM is regarded as a black-box and the UW addresses all uncertainty sources of the onion shell model [17]. The onion shell model classifies potential causes of uncertainties into model fit, input quality, and scope compliance [14]. Model-fit-related uncertainty is attributed to inherent limitations of a DDM, e.g., due to insufficient training data. Input-quality-related uncertainty is due to quality limitations on input data to which the DDM is applied. Scope-compliance-related uncertainty results from applying a DDM outside its intended scope of application [17].

Factors influencing the uncertainty related to input quality and scope compliance are modeled in the corresponding quality and scope model, respectively (cf. Fig. 1) [18].

In a *traffic sign recognition* (TSR) use case, *scope factors* such as the current GPS location could be used to assess whether a DDM is being used outside its *target application scope* (TAS), e.g., a limited spatial region such as Germany. To estimate the probability of the DDM being used outside its TAS, the scope factors are evaluated by a scope compliance model regarding the current situation. This can be based on fixed boundary checks or the computation of a similarity degree between the data at runtime and the data used during DDM development [18].

*Quality factors* (QF) could include rain or fog, which can obstruct visibility. In the quality impact model, a decision-tree-based approach is utilized, which considers the QFs for decomposing the TAS of the DDM into areas of similar uncertainties that can be safeguarded by taking into account a requested level of confidence. As decision trees exhibit transparency properties, this approach allows domain experts to verify the created quality impact model. Based on the uncertainties from both the quality impact model and the scope compliance model, a dependable uncertainty estimate is provided for an outcome of a DDM [11].

***Information/Data Fusion (IF)*** is defined by Wald as "a formal framework in which are expressed means and tools for the alliance of data originating from different sources. It aims at obtaining information of greater quality; the exact definition of 'greater quality' will depend upon the application." [19]

IF is most prominently applied for multimodal perception where it is also commonly known as *sensor fusion*. In the automotive domain, IF is used, for example, for multimodal object detection using information from LIDAR, RADAR, (stereo) RGB cameras, and further sensors [20] [21]. Often, end-to-end black-box models such as deep neural networks are applied to obtain fused outcomes [20].

IF can also be applied to combine the outcomes of *multiple ML models* [22]. For classification tasks, as, e.g., in the case of TSR, a simple fusion approach is majority voting [22] [13]. If the information sources have different reliabilities, their reliabilities can also be considered in the IF approach [12]. Empirical evidence shows, however, that there is no overall best combining rule [23].

IF over *multiple measurements* of a single source, e.g., one sensor, is another possibility to increase prediction accuracy. Here, results from multiple sensor readings are fused successively over time to arrive at an improved decision. IF over time can be applied, for example, for object recognition if the position of the object is tracked with sufficiently high accuracy to avoid ambiguities. Existing methods for traffic sign tracking

[24] [25] apply object detection and Kalman filtering to distinguish traffic signs using their position information.

***Uncertainty Fusion (UF)*** is applied in addition to IF to arrive at joint uncertainty estimates for fused model outcomes. Existing UF approaches [26] [27] often rely on the assumption of independence, although statistical dependencies between the occurrences of wrong DDM outcomes usually have to be expected.

A *naïve fusion* approach assumes the precondition that there are no dependencies between uncertainties $u_i$ obtained from different points in time $t_i$ or sources and define the joint uncertainty as their product:

$$u_n^{(naive)} = \prod_{i=0}^{n} u_i \quad (1)$$

To avoid the independence assumption, UF can also be performed by choosing the prediction with the lowest uncertainty among all predictions. This *opportune fusion* approach is a valid approach if the uncertainty estimates are never overconfident.

$$u_n^{(opportune)} = \min_{i \in \{0,\cdots,n\}} u_i \quad (2)$$

Usually, uncertainty can, however, only be estimated with high confidence and some low probability for overconfident estimates. Opportunistically choosing the lowest uncertainty prediction can thus cause an increase in overconfidence because overconfident predictions are chosen more often when using the opportune model.

On the other hand, a *worst-case fusion* approach combines uncertainties by taking the most conservative estimate among redundant perception components.

$$u_n^{(worstcase)} = \max_{i \in \{0,\cdots,n\}} u_i \quad (3)$$

This approach can help to reduce overconfidence. It is, however, also likely to be overly conservative. Overly conservative uncertainty predictions negate the advantages expected from using IF and can make the system unusable or at least result in performance restrictions.

### III. TIMESERIES-AWARE UNCERTAINTY WRAPPER

In this section, extensions to the existing UW framework are described. These extensions are used to make UW timeseries-aware. We give an overview of how a *timeseries-aware UW* (taUW) can use information from successive DDM *outcomes* (o) at runtime, traditional stateless *input quality factors* (QF), and *timeseries-aware quality factors* (taQF), which we specifically designed for timeseries information. Fig. 2 depicts the key concepts and architecture of the proposed taUW.

UWs based on the existing framework are stateless, i.e., their estimate $u_i$ is based only on the input during the corresponding timestep $t_i$. Therefore, the first part of the extension is a *timeseries buffer* that temporarily stores interim results during each timestep. The buffer is cleared at the onset of a new timeseries. A *timeseries* is defined as a series of consecutive inputs related to one measurement object. In the TSR use case we consider, the tracking component detects a new timeseries whenever the location of the detected object changes, i.e., the predictions might relate to a different traffic sign and thus also have a different ground truth.

Through the timeseries buffer, the *information fusion* component can access past and current DDM outcomes during each timestep. It fuses the DDM outcomes using an information fusion approach (infFuse) to achieve improved and timeseries-aware DDM predictions. In the following, we will refer to the ***fused outcome*** at timestep $t_i$ as

$$o_i^{(if)} = o^{(if)}(t_i) = \text{infFuse}(o_0, \ldots, o_i).$$

The other component that accesses the timeseries buffer is the *timeseries-aware quality model* (taQM) as part of the extended uncertainty wrapper. Using this quality model, taQF are derived from the *timeseries features* in the buffer, i.e., series of uncertainty estimates ($u_{j=0\ldots i}$) and DDM outcomes ($o_{j=0\ldots i}$), as well as the results of the stateless QF for the current timestep $t_i$. All these factors can then be used by the timeseries-aware *quality impact model* (taQIM) to make dependable uncertainty estimates. The timeseries-aware quality impact model is created similarly to the quality impact model of the uncertainty wrapper and retains its transparency property.

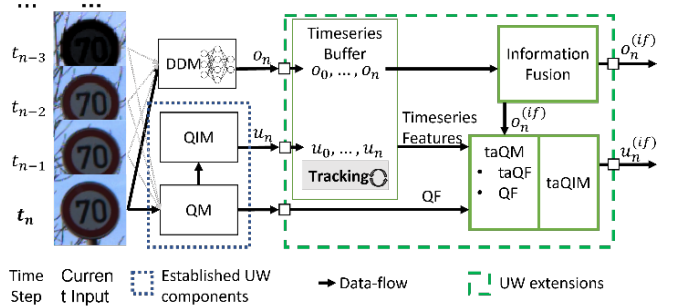

Fig. 2. Application of the timeseries-aware uncertainty wrapper architecture for traffic-sign recognition. During each timestep, classical uncertainty wrapper components are applied to the input and fed into a timeseries buffer. Using this buffer, timeseries-aware uncertainty wrapper components can make improved predictions and uncertainty estimations.

In this paper, we propose several timeseries-aware quality factors (taQF), which are described in the following. These factors are features that can be derived from semantic properties of the timeseries that are easily interpretable and are independent of the specific use case of TSR.

**taQF1** is the ***ratio*** of DDM outcomes in conformity with the current fused outcome $o_i^{(if)}$, $\frac{1}{i+1} \left| \left\{ j | o_j = o_i^{(if)} \right\}_{j=0\ldots i} \right|$

*Rationale.* The more often the current fused outcome was predicted in the timeseries compared to other outcomes, the more certainty there is about the current fused outcome.

**taQF2** is the ***length*** $i + 1$ of the current timeseries up to the current timestep $t_i$.

*Rationale.* The length is expected to influence the relevance of taQF in comparison to classical stateless QF. For longer timeseries, the *ratio*, e.g., is expected to be more meaningful and therefore more relevant for the uncertainty estimation.

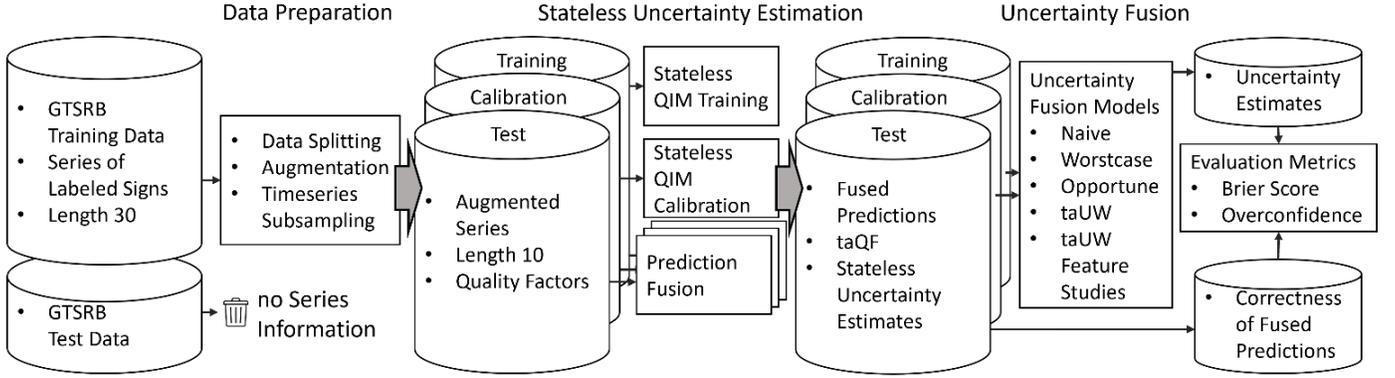

Fig. 3. Summary of study execution plan.

**taQF3** is the *size* of the set of unique outcomes in the current timeseries up to the current timestep $t_i$, i.e., $|\{o_j\}_{j=0...i}|$ can be used to estimate uncertainty.

*Rationale.* Higher variety in DDM outcomes can be an indicator of higher uncertainty.

**taQF4** is the *cumulative certainty* and defined as the sum of all certainties $c_{j=1...i}$ in the current timeseries up to the current timestep $t_i$. Previous outcomes that disagree with the current fused outcome are assumed to have a certainty of zero:

$$\sum_{j=0}^{i} c_j \text{ with } c_j = 1 - u_j \text{ if } o_j = o_i^{(if)} \text{ else } 0$$

*Rationale.* The *cumulative certainty* factor considers the uncertainty in previous timesteps in the estimate of the joint uncertainty. If we have more certainty in the current fused outcome, we can assume lower uncertainty.

Using these quality factors, it is possible to build a model-agnostic, transparent, and timeseries-aware uncertainty wrapper for dependable uncertainty estimates.

## IV. STUDY PLANNING AND EXECUTION

We conducted a study to evaluate the extensions to the uncertainty wrapper and the proposed timeseries-aware quality factors (taQF). The evaluation was performed in a traffic sign recognition task. This section explains how the study was conducted. In the first subsection, our research questions and hypotheses are presented while the study design is described in the second subsection. The third subsection presents details of the implementation and execution of the study.

### A. Research Questions

Aiming to evaluate timeseries-aware uncertainty wrappers, we investigated the following research questions during our study using the application of traffic sign recognition based on timeseries data:

**RQ1** Can prediction accuracy be improved using information fusion over time and is it effectively applicable even for shorter timeseries?

**RQ2** How does the timeseries-aware uncertainty wrapper compare to **(a)** the basic UW and **(b)** to the naïve, the opportune, and the worst-case uncertainty fusion (UF) approaches?

**RQ3** Which are the most relevant quality factors for timeseries-aware UW to improve uncertainty estimation performance?

Although it is possible in theory that conflicting information from individual timesteps will lead to a reduction in the quality of a fused prediction outcome, we expect an increase in the accuracy of fused outcomes and an increase in quality for timeseries-aware uncertainty estimates. Furthermore, we expect that the consideration of each of the taQF will lead to improvement of the uncertainty predictions.

### B. Study Design

In this subsection, we describe the study design in terms of the context, including the examined prediction task, the data used for the evaluation, and the evaluation metrics we used to score the examined uncertainty estimators. An overview of the study execution is illustrated in Fig. 3.

#### 1) Context

The *task* of the DDM examined during this study is traffic sign recognition, i.e., correctly classifying traffic signs on given images. We assume the target application scope (TAS) of the DDM as a roadworthy passenger car. The car is driving within Germany at different points in time, hence, being confronted with various weather conditions and operation conditions (e.g., a steamed-up camera lens).

#### 2) Dataset

For our study, different datasets were needed with information on their timeseries: a training dataset to build an uncertainty estimator; a calibration dataset to derive unbiased uncertainty estimates from unseen data; and a test dataset that is representative for the TAS to evaluate the uncertainty estimator.

The datasets for the considered use case are based on the German Traffic Sign Recognition Benchmark (GTSRB) dataset [28], with 1307 series of traffic sign images as the car approaches the traffic sign in its training dataset containing in total 39,209 images. The series comprise between 29 and 30 images each. The test dataset of GTSRB does not contain timeseries data and was not used further. As the GTSRB data does not provide further information on the contextual setting, which might influence the uncertainty (e.g., weather conditions), we augmented the images with 9 types of quality deficits (i.e., rain, darkness, haze, natural backlight, artificial backlight, dirt on the traffic sign, dirt on the sensor lens, steamed-up sensor lens, and motion blur) based on a given realistic situation setting. The situation settings were generated based on historical weather data from Deutscher Wetterdienst [29] and street locations within the target application scope from OpenStreetMap [30]. To this end, the data augmentation framework by Jöckel and Kläs [31] was applied with slight variations to consider the nature of timeseries data. Each series of images showing the same physical traffic sign was assigned one situation setting with its associated quality deficits, which were then propagated through the series. The setting and most of the quality deficits did not vary throughout the series, except for motion blur and artificial backlight, for which the conditions might change within the series.

The available 1307 timeseries were randomly split into datasets, resulting in a training dataset with 522 series (15,660 images), a calibration dataset with 392 series (11,760 images), and a test dataset with 392 series (11,760 images). The training data was augmented for each quality deficit with low, medium, and high intensity, resulting in an augmented training dataset with 409,440 images. The calibration and test datasets were augmented by selecting random situation settings out of a set of 2.7 million realistic settings that emulate the target application scope, leading to representative datasets. Each original series was augmented 28 times (each time based on a different setting) to keep the original ratio with the training data, resulting in calibration and test data with 329,280 images. To avoid biased uncertainty predictions due to the distance from the traffic signs, the timeseries in the calibration and test set were subsampled, i.e., from each timeseries a subseries of length 10 was chosen with a uniformly random starting time step.

*3) Evaluation of Uncertainty Estimates*

Based on the test dataset, we evaluated all uncertainty wrapper approaches. Using the Brier score ($bs$), a measure of the mean squared difference between the probability of an outcome and the actual outcome, the performance of the uncertainty estimates was assessed [32]. Variance ($var$), resolution ($res$), and unreliability ($unr$) allow decomposition with $bs = var - res + unr$ [33]. Here, $var$ is invariant to the uncertainty estimation approach, as it only depends on the error rate of the DDM. If the error rate of a DDM is high, this is reflected in a high $var$ value. The extent to which the case-specific uncertainty estimates deviate from the overall uncertainty is described by $res$. Since higher $res$ values are better and are bounded by $var$, we use $unspecificity$, which derives from $var - res$. The last component is $unr$, which describes how well the estimated uncertainty is calibrated to the observed error rate of the DDM. A smaller value indicates lower unreliability, i.e., good calibration. Furthermore, with $overconfidence$ we measure the portion of $unr$ due to uncertainty estimates that underestimate the observed error rate. The residual portion of $unr$ is referred to as $underconfidence$ [18].

*C. Study Execution*

In this subsection, we present the neural network architecture and the training of the DDM, the information fusion approach employed for the outcomes, and the calibration of the timeseries-aware uncertainty wrapper.

*1) DDM Training*

For the DDM, we used a convolutional neural network (CNN) with a model architecture roughly based on [34] in a variant built in [18]. The DDM was trained on an augmented training dataset based on the GTSRB training dataset without considering timeseries. The accuracy of the complete augmented timeseries-aware test dataset was 91.7%.

*2) Uncertainty Wrapper Calibration*

Building an uncertainty wrapper requires training a decision tree as the quality impact model (QIM). This decision tree was optimized using the CART algorithm based on the *gini index* as an approximation for entropy using the training data. The training was conducted up to a maximum depth of 8 without pruning during this phase. Afterwards, the quality impact model was calibrated on the subsampled calibration dataset. For the calibration, all leaves were pruned so that each leaf in the decision tree was left with at least 200 samples. Then statistical uncertainty guarantees were calculated for each leaf at a confidence level of 0.999. Our investigation focused on uncertainty related to timeseries-aware input quality. Therefore, all datapoints were chosen to be within the target application scope and the construction of a scope compliance model was omitted. Subsequently, the timeseries-aware quality impact model was trained using the same procedure with the proposed taQF in addition to the stateless QF of the current input. The calibration was conducted using the length 10 timeseries from the calibration dataset.

*3) Information Fusion*

We applied information fusion over time to increase the prediction accuracy of the classifications. Majority voting is a simple but common IF approach to combine different outcomes for a classification task [22]. It was applied as a transparent method to aggregate the iterative prediction outcomes $o_{j=1...i}$. In this approach, the mode of the number of momentaneous predictions per class is chosen as the fused outcome $o_i^{(if)}$. To resolve ties, the most recent momentaneous prediction is chosen in case two or more classes were predicted the greatest number of times. We applied the same IF approach to evaluate all UF models.

V. STUDY RESULTS AND DISCUSSION

The results of the study are presented in this section. Each subsection is dedicated to one of the research questions. The results for each research question are discussed after the presentation of the empirical evaluation.

**RQ1** To evaluate the timeseries-aware uncertainty wrapper, majority voting as a simple IF approach was applied to the DDM predictions. Fig. 4 shows the results itemized by the position within the timeseries. For isolated as well as IF-based outcomes, the misclassification rate decreases over the timeseries since we consider consecutive timesteps in a driving situation, meaning the pixel size of the traffic sign image increases, which generally reduces the misclassification rate. During the first two steps, DDM +IF and isolated DDM prediction outcomes coincide and thus result in the same misclassification rate. With three or more timesteps, the fused predictions are more accurate than the isolated predictions. The DDM itself has a misclassification rate of 7.89% on the images of the length 10 timeseries. Using *majority voting* as IF approach, it was possible to reduce the misclassification rate to 5.57% on average over all timesteps, achieving 3.69% in timestep 10.

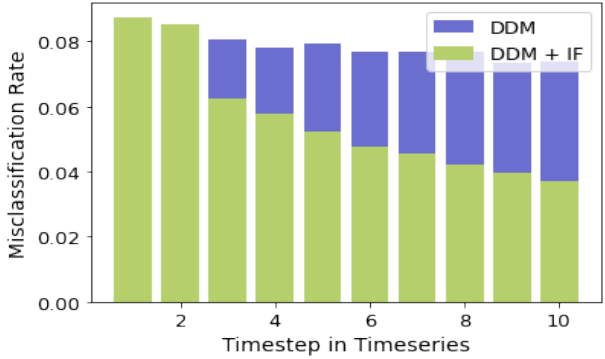

Fig. 4. Misclassification rate over several timesteps for *isolated* predictions (blue) and *information fusion* (green) based on majority voting.

*Discussion:* Despite its simplicity, the employed IF approach turned out to be quite effective. The improvement strongly depends on the number of images from the same object. Even after ten images, the improvement in accuracy does not appear to reach saturation. Thus, with longer timeseries, an even better result could be achieved.

Although the experiments were only performed on natural data that was artificially augmented, we believe that the results are generalizable for natural data because of the high degree of realism of the augmentations and reports of improvements of a similar magnitude in the literature [22] [20]. The majority vote IF approach could likely be improved further by employing more complicated deep neural networks at the cost of transparency.

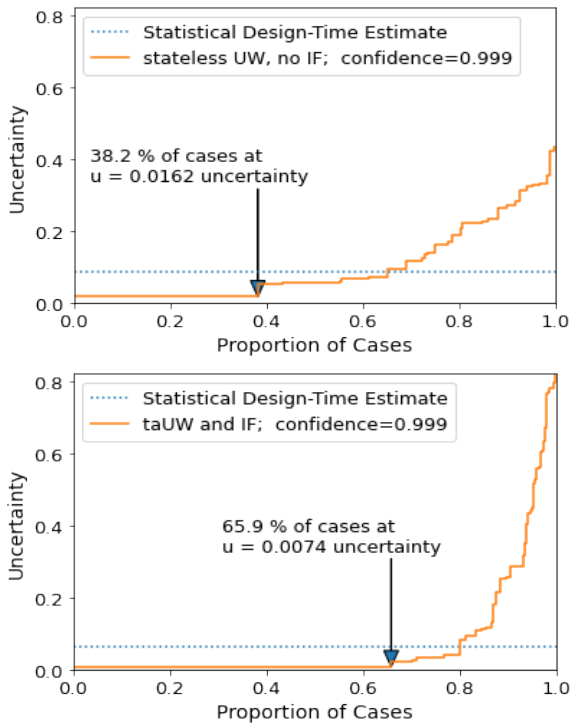

Fig. 5. Distribution of uncertainty across cases illustrated for the classical stateless UW (top) and the proposed timeseries-aware uncertainty wrapper in combination with information fusion (bottom). The proportion of cases for which the lowest uncertainty can be predicted with 99.9% confidence is highlighted by the arrow.

**RQ2 (a)** To evaluate the uncertainty estimates, we examined several aspects. Fig. 5 shows a comparison of the uncertainty distribution between the proposed timeseries-aware uncertainty wrapper (taUW) in combination with IF and traditional stateless uncertainty wrapper as a baseline. Using the proposed approach, the lowest uncertainty of $u = 0.0072$ can be guaranteed for 65.9% of the cases with 99.9% confidence.

To compare the quality of the uncertainty estimates, the Brier loss score, its components, and an overconfidence metric were used as metrics. In TABLE I. the evaluation results are shown. The proposed taUW in combination with IF achieved the best scores for all evaluation metrics. The applied majority vote IF approach reduced the variance component of the Brier score compared to isolated, momentaneous predictions.

*Discussion:* The results from Fig. 5 show that UF over time is especially beneficial for situations with low uncertainty because even though the DDM only has an accuracy of 92.1% on the images of the length 10 timeseries, predictions can be made with more than 99% certainty in the majority of cases. By using the proposed model, the number of cases for which the lowest uncertainty can be guaranteed almost doubles while the amount of uncertainty that needs to be tolerated is reduced by more than half.

There are several conclusions that can be drawn from the results shown in TABLE I. First, we see that information fusion alone already leads to improved uncertainty estimates in terms of the Brier score and in terms of the Brier score components variance and unspecificity. These reductions are expected because they are a direct result of higher accuracy of the fused predictions. On the other hand, we see a slight worsening of unreliability and overconfidence. This is likely due to the relative reduction in the predictive power of the uncertainty wrapper compared to the improved prediction model. The results are promising for the development of real-world applications especially because unlike sensor fusion, IF over time does not require additional sensor hardware.

TABLE I. EVALUATION OF DIFFERENT UNCERTAINTY MODELS

| Approach | Brier Loss Score and its Components | | | | |
|---|---|---|---|---|---|
| | **Brier Score** | **Variance** | **Unspecificity** | **Unreliability** | **Over-confidence** |
| **Stateless UW no IF + no UF** | 0.0661 | 0.0726 | 0.0651 | 0.00094 | 7.0e-06 |
| **(Fused) IF + no UF** | 0.0498 | **0.0526** | 0.0487 | 0.00112 | 3.9e-05 |
| **IF + Naïve UF** | 0.0490 | **0.0526** | 0.0434 | 0.00565 | 5.6e-03 |
| **IF + Worst-case UF** | 0.0588 | **0.0526** | 0.0488 | 0.01002 | 5.1e-07 |
| **IF + Opportune UF** | 0.0481 | **0.0526** | 0.0466 | 0.00152 | 1.8e-04 |
| **IF + taUW** | **0.0356** | **0.0526** | **0.0346** | **0.00101** | **0.0** |

Brier scores of the evaluated fusion models and the Brier score components *variance, unspecificity,* and *calibration* as well as the *overconfidence* of the uncertainty predictions. The evaluated conditions describe whether information fusion was applied instead of considering only isolated predictions and which uncertainty fusion (UF) model was used. The condition is timeseries-unaware if no UF was used. The best (**lowest**) scores are highlighted in bold.

**RQ2 (b)** To answer the second part of the research question, we compared the calibration plots of different UF approaches. The calibration of the uncertainty estimation models can be examined in detail in Fig. 6. Quantiles below the thin dotted line are overconfident, i.e., the model is wrong, and the actual uncertainty is underestimated. Conversely, quantiles above this line are underconfident. We can see that the naïve UF approach is highly overconfident. The worst-case approach is the most conservative. Both the opportune model and the proposed timeseries-aware uncertainty wrapper appear to be well calibrated. The range of different uncertainty predictions of the taUW is the largest of all considered uncertainty estimation models.

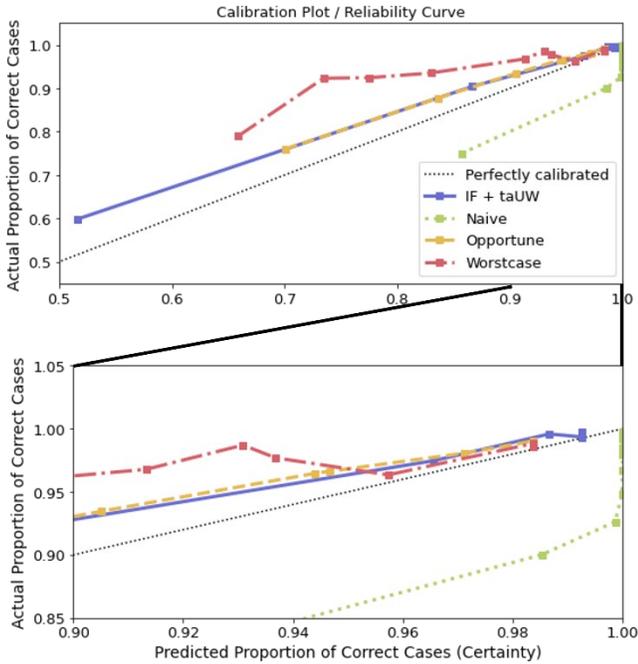

Fig. 6. Calibration plot illustrating the over- and underconfidence of different models for uncertainty estimation. Quantiles of the predicted certainty (1-uncertainty) are plotted against their actual correctness in 10% steps. Points above the thin dotted line indicate underconfidence. Points below that line indicate overconfidence.

*Discussion:* We compared the examined UF approaches and found that the worst-case UF approach performs the worst in terms of the Brier score. However, it has very low overconfidence, which means that the uncertainty estimates are dependable if the overestimation of uncertainties is uncritical because the approach is overly conservative. In most cases, the worst-case model overestimates the actual uncertainty. This can also be seen from the calibration plot in Fig. 6. Overly conservative uncertainty estimates can lead to a reduction in performance during practical applications due to restrictive safety margins. By contrast, the naïve model has a better Brier score but much higher overconfidence.

Almost all quantiles of the calibration curve are overconfident. This is unsurprising because the DDM predictions violate the naïve independence assumption. DDM errors show strong dependencies within the steps of a timeseries because constant or slowly changing environment factors lead to systematic mistakes and thus it cannot be assumed that successive DDM misclassifications will occur purely at random. The result shows that the Brier score should not be considered in isolation when evaluating uncertainty predictions, and it exemplifies that a naïve independence assumption can be quite inappropriate in practice.

The opportune UF model and the proposed taUW are close to each other in the calibration plot. A closer look shows that the taUW has a larger range of predicted uncertainties. This is also expressed in the lower unspecificity in TABLE I. This result can be explained by the use of timeseries-aware features that allow the timeseries-aware quality impact model to better distinguish between clusters of input quality deficits and thus arrive at more fine-grained and adaptive uncertainty estimates.

Furthermore, the results show that the opportune model was more overconfident than all other models except the naïve model. This is to be expected because the opportune model is more likely to base its uncertainty estimate on overconfident model outcomes, as described in Section II.

**RQ3** To test the impact of each of the timeseries-aware features on the Brier score, we conducted a feature importance study. The results of this study are shown in Fig. 7. Generally, the Brier score improves when more features are used. The best result can already be achieved using only the *ratio* and the *certainty* feature, indicating a certain degree of redundancy among the proposed features.

| **Feature Importance** | | R**LSC** 0.0365 | | |
|---|---|---|---|---|
| Brier Score | | RLS**C** 0.0457 | RL**SC** 0.0361 | R**LSC** 0.0358 |
| | | RL**SC** 0.0431 | R**LSC** 0.0387 | R**LSC** 0.0356 |
| RLSC 0.0498 | R**LSC** 0.0498 | R**LSC** 0.0356 | R**LSC** 0.0356 | **RLSC** 0.0356 |
| | R**LSC** **0.0370** | R**LSC** 0.0372 | R**LSC** 0.0362 | |
| | | R**LSC** 0.0362 | | |

Features Used Yes/No
- **R**atio
- **L**ength
- **S**ize
- **C**ertainty

Fig. 7. Results of a feature importance study. In each column, the Brier loss score of the uncertainty predictions using combinations of a number of timeseries-aware features is shown.

*Discussion:* The *ratio* and the *certainty* feature appear to be the most important features. These features have good predictive power when used individually and achieve the optimal result with the fewest number of features when used in combination. The *length* of the current timeseries feature does not lead to any improvement of the Brier score. However, in combination with one other feature, the *length* feature leads to improved uncertainty estimates. The *size* feature is the second-best feature when used on its own. Like the *length* feature, its use in combination with *certainty* and *ratio* does not lead to a better Brier score. Due to the redundancy in the feature set, it is likely that not all features are required. Experiments on other datasets are required to determine whether the results are stable and whether there is an overall best set of timeseries-aware features.

## VI. Conclusion

In this paper, we have proposed an extension to the uncertainty wrapper framework that makes it timeseries-aware, enabling improved uncertainty predictions in systems that use artificial intelligence. We demonstrated that it is possible to significantly reduce misclassification rates by applying information fusion over time. We compared the performance of our timeseries-aware uncertainty wrapper to already proposed uncertainty fusion approaches, including naïve, opportune, and worst-case. In our study, the timeseries-aware uncertainty wrapper outperformed these approaches in terms of specificity, reliability, and avoiding overconfidence. It also provided the best overall uncertainty quantification performance as measured by the Brier score. Additionally, we conducted a feature importance study to gain insights into which timeseries-aware features contribute most significantly to uncertainty estimation.

Overall, our work highlights the importance of considering timeseries data in uncertainty prediction and we hope to provide a valuable contribution to runtime V&V and the development of more dependable AI systems.

## Acknowledgment

Parts of this work have been funded by the project "LOPAAS" as part of the internal funding program "ICON" of the Fraunhofer-Gesellschaft. We would like to thank Sonnhild Namingha and Thorsten Honroth for the initial paper review.

## References

[1] A. I. Abiodun, A. Jantan, A. E. Omolara, K. V. Dada, N. A. E. Mohamed and H. Arshad, "State-of-the-art in artificial neural network applications: A survey," *Heliyon,* vol. 4, no. 11, p. e00938, 2018.

[2] M. Kläs, "Towards Identifying and Managing Sources of Uncertainty in AI and Machine Learning Models - An Overview," 2018. [Online]. Available: https://arxiv.org/abs/1811.11669.

[3] R. Salay, R. Queiroz and K. Czarnecki, *An Analysis of ISO 26262: Using Machine Learning Safely in Automotive Software,* 2017.

[4] M. Kläs, T. Bauer, A. Dereani, T. Söderqvist and P. Helle, "A large-scale technology evaluation study: effects of model-based analysis and testing," in *ICSE*, Florence, Italy, 2015.

[5] P. Helle, W. Schamai and C. Strobel, "Testing of autonomous systems– Challenges and current state-of-the-art," in *INCOSE*, Edinburg, Scotland, 2016.

[6] V. Riccio, G. Jahangirova, A. Stocco, N. Humbatova, M. Weiss and P. Tonella, "Testing machine learning based systems: a systematic mapping," *Empirical Software Engineering,* vol. 25, 2020.

[7] L. Jöckel, T. Bauer, M. Kläs and M. Hauer, "Towards a Common Testing Terminology for Software Engineering and Data Science Experts," in *PROFES*, Turin, Italy, 2021.

[8] J. G. Rivera, A. A. Danylyszyn, C. B. Weinstock, L. R. Sha and M. J. Gagliardi, "An Architectural Description of the Simplex Architecture," Software Engineering Institute, Carnegie Mellon University, Pittsburgh, Pennsylvania, 1996.

[9] D. T. Phan, R. Grosu, N. Jansen, N. Paoletti, S. A. Smolka and S. D. Stoller, "Neural Simplex Architecture," in *NFM*, CA, USA, 2020.

[10] J. Groß, R. Adler, M. Kläs, J. Reich, L. Jöckel and R. Gansch, "Architectural patterns for handling runtime uncertainty of data-driven models in safety-critical perception," in *SafeComp*, Munich, 2022.

[11] M. Kläs and L. Sembach, "Uncertainty wrappers for data-driven models – Increase the transparency of AI/ML-based models through enrichment with dependable situation-aware uncertainty estimates," in *WAISE*, Turku, Finland, 2019.

[12] G. L. Rogova and V. Nimier, "Reliability In Information Fusion: Literature Survey," in *Fusion*, Stockholm, Sweden, 2004.

[13] J. Kittler, M. Hatef, R. P. Duin and J. Matas, "On combining classifiers," *IEEE transactions on pattern analysis and machine intelligence,* vol. 20, no. 3, pp. 226-239, 1998.

[14] M. Kläs and A. M. Vollmer, "Uncertainty in Machine Learning Applications – A Practice-Driven Classification of Uncertainty," in *WAISE*, 2018.

[15] M. Kläs, R. Adler, I. Sorokos, L. Jöckel and J. Reich, "Handling Uncertainties of Data-Driven Models in Compliance with Safety Constraints for Autonomous Behaviour," in *EDCC*, 2021.

[16] T. Bandyszak, L. Jöckel, M. Kläs, S. Törsleff, B. Weyer and B. Wirtz, "Handling Uncertainty in Collaborative Embedded Systems Engineering," in *Model-Based Engineering of Collaborative Embedded Systems*, Springer, 2021, pp. 147-170.

[17] M. Kläs and L. Jöckel, "A Framework for Building Uncertainty Wrappers for AI/ML-based Data-Driven Components," in *WAISE*, Lisbon, Portugal, 2020.

[18] L. Jöckel and M. Kläs, "Could We Relieve AI/ML Models of the Responsibility of Providing Dependable Uncertainty Estimates? A Study on Outside-Model Uncertainty Estimates," in *SafeComp*, York, United Kingdom, 2021.

[19] L. Wald, "Some terms of reference in data fusion," *IEEE Transactions on geoscience and remote sensing,* vol. 37, no. 3, pp. 1190-1193, 1999.

[20] M. Bijelic, T. Gruber, F. Mannan, F. Kraus, W. Ritter, K. Dietmayer and F. Heide, "Seeing Through Fog Without Seeing Fog: Deep Multimodal Sensor Fusion in Unseen Adverse Weather," in *CVPR*, online, 2020.

[21] D. J. Yeong, G. Velasco-Hernandez, J. Barry and J. Walsh, "Sensor and Sensor Fusion Technology in Autonomous Vehicles: A Review," *Sensors,* vol. 21, no. 6, p. 2140, 2021.

[22] G. Rogova, "Combining the results of several neural network classifiers," in *Classic Works of the Dempster-Shafer Theory of Belief Functions*, Berlin, Heidelberg, Springer, 2008, pp. 683-692.

[23] R. P. W. Duin and D. M. J. Tax, "Experiments with classifier combining rules," *Multiple classifier systems,* vol. 1857, pp. 16-29, 2000.

[24] C.-Y. Fang, S.-W. Chen and C.-S. Fuh, "Road-sign detection and tracking," *IEEE transactions on vehicular technology,* vol. 52, no. 5, pp. 1329-1341, 2003.

[25] A. Gudigar, S. Chokkadi and R. U, "A review on automatic detection and recognition of traffic sign," *Multimedia Tools and Applications,* vol. 75, pp. 333-364, 2016.

[26] I. Kurzidem, A. Saad and P. Schleiss, "A Systematic Approach to Analyzing Perception Architectures in Autonomous Vehicles," in, *IMBSA*, Lisbon, Portugal, 2020.

[27] P. Schleiss, Y. Hagiwara, I. Kurzidem and F. Carella, "Towards the Quantitative Verification of Deep Learning for Safe Perception," in *ISSREW*, Charlotte, North Carolina, USA, 2022.

[28] "German Traffic Sign Benchmarks," [Online]. Available: http://benchmark.ini.rub.de/?section=gtsrb. [Accessed 13 11 2020].

[29] "Climate Data Center," 27 03 2023. [Online]. Available: https://cdc.dwd.de/portal/.

[30] "OpenStreetMap," 13 11 2020. [Online]. Available: https://www.openstreetmap.de/.

[31] L. Jöckel and M. Kläs, "Increasing trust in data-driven model validation – A framework for probabilistic augmentation of images and meta-data generation using application scope characteristics," in *SafeComp*, Turku, Finland, 2019.

[32] G. W. Brier, "Verification of Forecasts Expressed in Terms of Probability," *Monthly Weather Review,* vol. 78, no. 1, pp. 1-3, 1950.

[33] A. H. Murphy, "A new vector partition of the probability score," *Journal of Applied Meteorology,* vol. 12, no. 4, p. 595–600, 1973.

[34] Á. Arcos-García, J. Alvarez-Garcia and L. Soria Morillo, "Deep neural network for traffic sign recognition systems: An analysis of spatial transformers and stochastic optimisation methods, *Neural Networks* vol. 99, 2018.

[35] L. Jöckel, M. Kläs and S. Martínez-Fernández, "Safe Traffic Sign Recognition through Data Augmentation for Autonomous Vehicles Software," in *QRS-C*, Sofia, Bulgaria, 2019.